%% file: few_labels.tex
\documentclass{article}
\pdfoutput=1
\usepackage{spconf,amsmath,graphicx}

\usepackage{microtype}

\usepackage{booktabs} % for professional tables
\usepackage{tikz}
\usepackage{hyperref,graphicx,amsmath,amsfonts,subcaption,amssymb,bm,url,breakurl,epsfig,epsf,color,MnSymbol,mathbbol,fmtcount,semtrans,caption,comment}

\usepackage{color}
\usepackage[utf8]{inputenc} % allow utf-8 input
\usepackage[T1]{fontenc}    % use 8-bit T1 fonts
\usepackage{hyperref}       % hyperlinks
\usepackage{url}            % simple URL typesetting

\usepackage{nicefrac}       % compact symbols for 1/2, etc.
\newcommand{\som}[1]{}%\marginpar{\color{darkblue}\tiny\ttfamily SO: #1}}
\newcommand{\so}[1]{{\color{black}#1}}

\input{notation}

\usepackage[ruled,vlined]{algorithm2e}

\title{On the Marginal Benefit of Active Learning:\\Does Self-Supervision Eat Its Cake?}

\name{Yao-Chun Chan$^{\star\alpha}$ \qquad Mingchen Li$^{\star\alpha}$ \qquad Samet Oymak$^{\dagger}$ \thanks{$^{\alpha}$ Equal contribution}}
\address{$^{\star}$ Department of Computer Science and Engineering, University of California, Riverside\\
$^{\dagger}$ Department of Electrical and Computer Engineering, University of California, Riverside}
\begin{document}
	%\title{A Simple Framework for Active Semi-supervised Learning of Self-supervised Pretraining}

	\maketitle
	\input{intro}

\input{problem}
	\input{approach}

	\input{experiments}

\input{related}
	\input{conclusion}

	\newpage
	\clearpage
	\bibliography{bibfile}
	\bibliographystyle{IEEEbib}

\end{document}

%% file: notation.tex
\usepackage{hyperref,graphicx,amsmath,amsfonts,amssymb,bm,url,breakurl,epsfig,epsf,color,MnSymbol,mathbbol,fmtcount,semtrans,caption,subcaption,multirow,comment, boldline}
\usepackage[noend]{algpseudocode}
\usepackage{wrapfig}
\usepackage{amssymb}

\usepackage[utf8]{inputenc} % allow utf-8 input
\usepackage[T1]{fontenc}    % use 8-bit T1 fonts
\usepackage{url}            % simple URL typesetting
\usepackage{booktabs}       % professional-quality tables
\usepackage{amsfonts}       % blackboard math symbols
\usepackage{nicefrac}       % compact symbols for 1/2, etc.
\usepackage{microtype}      % microtypography
\usepackage{mathtools}

\definecolor{darkred}{RGB}{150,0,0}
\definecolor{darkgreen}{RGB}{0,150,0}
\definecolor{darkblue}{RGB}{0,0,200}
\hypersetup{colorlinks=true, linkcolor=darkred, citecolor=darkgreen, urlcolor=darkblue}

\newcommand{\beq}{\begin{equation}}

\newcommand{\eeq}{\end{equation}}

\definecolor{emmanuel}{RGB}{255,127,0}

\numberwithin{equation}{section} 

\def \endprf{\hfill {\vrule height6pt width6pt depth0pt}\medskip}

%% file: intro.tex
\begin{abstract}
	Active learning is the set of techniques for intelligently labeling large unlabeled datasets to reduce the labeling effort. In parallel, recent developments in self-supervised and semi-supervised learning (S4L) provide powerful techniques, based on data-augmentation, contrastive learning, and self-training, that enable superior utilization of unlabeled data which led to a significant reduction in required labeling in the standard machine learning benchmarks. A natural question is whether these paradigms can be unified to obtain superior results. To this aim, this paper provides a novel algorithmic framework integrating self-supervised pretraining, active learning, and consistency-regularized self-training. We conduct extensive experiments with our framework on CIFAR10 and CIFAR100 datasets. These experiments enable us to isolate and assess the benefits of individual components which are evaluated using state-of-the-art methods (e.g.~Core-Set, VAAL, simCLR, FixMatch). Our experiments reveal two key insights: (i) Self-supervised pre-training significantly improves semi-supervised learning, especially in the few-label regime, (ii) The benefit of active learning is undermined and subsumed by S4L techniques. Specifically, we fail to observe any additional benefit of state-of-the-art active learning algorithms when combined with state-of-the-art S4L techniques. 
	
	% (pretraining, active learning, and self-training)
	% 	 provide a strong support on multiple mainstream active learning and self/semi-supervised learning
	%	 which benefits from unlabeled data using contrastive learning and consistency regularization achieve far better performance than state-of-the-art active learning algorithms with fewer labeled data.
	
	%Active learning is a useful approach for data selection when training with large unlabeled datasets. However, recent successes in self-supervised and semi-supervised learning which take benefit from unlabeled data using contrastive learning and consistency regularization achieve far better performance than state-of-the-art active learning algorithms with fewer labeled data. In this paper, we investigate the benefit of active learning on self-supervised and semi-supervised learning and provide two insights: (i) self-supervised pre-train significantly improves semi-supervised learning especially in fewer labeled setup, (ii) the benefit of active learning is undermined by self/semi-supervised learning with consistency regularization. Our experiments on CIFAR10 and CIFAR100 provide a strong support on multiple mainstream active learning and self/semi-supervised learning algorithms (Core-Set,VAAL,simCLR,MoCo,FixMatch).
\end{abstract}
\section{Introduction}
	Traditional supervised learning requires a large dataset with good-quality annotations. In practice, such datasets are often not available since annotation might be difficult, expensive and time consuming. In many real-world applications, such as medical imaging and activity recognition \cite{wang2019deep}, labeling becomes a significant challenge either because the dataset is very large or requires domain expertise. In comparison, access to unlabeled data can be relatively easy. Thus, the last decade witnessed a significant effort towards developing new methodologies to better utilize unlabeled data via transfer learning~\cite{tan2018survey}, semi-supervised learning~\cite{sohn2020fixmatch,berthelot2019remixmatch,smith2020building}, self-supervised learning~\cite{chen2020big,he2020momentum,chuang2020debiased} and active learning~\cite{sener2017active,sinha2019variational,mottaghi2019adversarial,kim2020task}.
	
	%Especially on the self-supervised pre-train model or under semi-supervised setup. 
	\som{What is pool-based?}
	Contemporary active learning approaches, especially the pool-based active learning methods~\cite{sener2017active,sinha2019variational,kim2020task}, \so{treat the unlabeled data as a pool to select the most informative samples and train the model only with the selected labeled samples.} Although \cite{mottaghi2019adversarial} proposes an approach which uses both active and semi-supervised learning to better utilize unlabeled data, it still requires a large number of labeled samples at the beginning and at each active learning epoch. All of these active learning approaches (\cite{sener2017active,sinha2019variational,kim2020task,mottaghi2019adversarial}) often require thousands of samples for labeling to achieve fully-supervised performance.
	
	In contrast, recent semi-supervised learning methods, such as FixMatch~\cite{sohn2020fixmatch}, UDA~\cite{xie2019unsupervised} and BOSS~\cite{smith2020building}, can accomplish comparable performance using 10 to 100 times fewer randomly selected labeled data by employing strong data-augmentation on unlabeled examples. Similarly, contrastive loss based self-supervised pretraining~\cite{chen2020simple,he2020momentum,chuang2020debiased} achieves fully-supervised performance when fine-tuning the pre-trained model with 10\% labeled data~\cite{chen2020big} which also outperforms traditional active learning. Thus, the success of self-/semi-supervised learning (S4L) raises the question that \emph{to what extent active learning is helpful in machine learning when combined with S4L}. \emph{Can we combine active learning with S4L techniques to further reduce the labeling effort?}
	%improve the semi-supervised learning when selecting labeled samples?}
		\begin{table*}[]
		\centering
		\resizebox{\textwidth}{!}{%
			\begin{tabular}{@{}ccccccccc@{}}
				\toprule
				& \multicolumn{4}{c}{Cifar 10} & \multicolumn{4}{c}{Cifar 100} \\ 
				\midrule
				\multicolumn{1}{c}{Methods} &40 Labels & 150 Labels & 250 Labels& Supervised & \multicolumn{1}{|c} {400} Labels & 1500 Labels & 2500 Labels & Supervised \\ 
				\midrule
				\multicolumn{1}{c|}{DCL + Random sampling + Pseudo-labeling}  & 70.77 & 75.11 & 78.83 & 94.67  & \multicolumn{1}{|c}{} 28.62 & 43.24 & 49.93  & 76.87  \\
				%\multicolumn{1}{c|}{MoCo + Random sampling + Pseudo-labeling} &  &  &  &  & \multicolumn{1}{|c}{}  &  &   &  \\
				\multicolumn{1}{c|}{DCL + Random sampling + FixMatch} & 93.23 & 93.90 & 94.90 & 94.67 & \multicolumn{1}{|c}{} 47.66 & 65.25 &  67.49 & 76.87  \\
				%\multicolumn{1}{c|}{MoCo + Random sampling + FixMatch} &  &  &  &  & \multicolumn{1}{|c}{}  &  &   &  \\
				\midrule
				\multicolumn{1}{c|}{DCL + Core-Set + Pseudo-labeling} & 65.01 & 75.88  & 80.75 & 94.67 & \multicolumn{1}{|c}{}  29.11 & 43.25 & 49.77  & 76.87  \\
				%\multicolumn{1}{c|}{MoCo + Core-Set + Pseudo-labeling} &  &  &  &  & \multicolumn{1}{|c}{}  &  &   &  \\
				%\multicolumn{1}{c|}{MoCo + VAAL + Pseudo-labeling}  &  &  &  &  & \multicolumn{1}{|c}{}  &  &   &  \\
				\multicolumn{1}{c|}{DCL + Core-Set + FixMatch}  &  93.21 &  92.76 &  94.53 & 94.67 & \multicolumn{1}{|c}{} 47.81 & 65.11  &   66.09 & 76.87  \\
				
				\multicolumn{1}{c|}{DCL + VAAL + Pseudo-labeling}  & 64.72  & 76.59  & 79.71  & 94.67 & \multicolumn{1}{|c}{} 28.54 & 44.02 &  50.07 & 76.87  \\
				\multicolumn{1}{c|}{DCL + VAAL + FixMatch}  & 93.37 & 93.84 &  94.61  & 94.67 & \multicolumn{1}{|c}{} 45.72 & 65.32 &    66.12 & 76.87  \\
				%\multicolumn{1}{c|}{MoCo + Core-Set + FixMatch} &  &  &  &  & \multicolumn{1}{|c}{}  &  &   &  \\
				%\multicolumn{1}{c|}{MoCo + VAAL + FixMatch}  &  &  &  &  & \multicolumn{1}{|c}{}  &  &   &  \\
				\bottomrule
			\end{tabular}
		}\caption{This table displays the performance of two key active learning methods (Core-Set and VAAL) and random sampling when combined with S4L techniques for different labeling budgets.}\label{table:ablation_active}\vspace{-13pt}
	\end{table*}
	 \begin{figure}[t]
	 	\centering
	 	\begin{tikzpicture}
	 	\node at (0,0) {\includegraphics[scale=1]{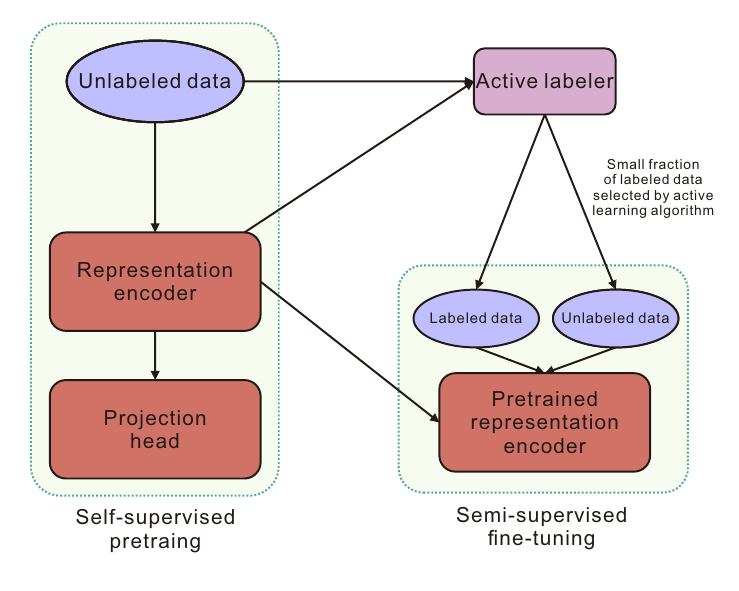}};
	 	\end{tikzpicture}\caption{\small{Proposed framework for combining active learning and self-/semi-supervised learning. We first obtain a pre-trained model from self-supervised learning such as simCLR~\cite{chen2020simple} or MoCo~\cite{he2020momentum}. Then the pre-trained model is used by an active labeler which selects most informative examples to label. The semi-supervised algorithm uses both labeled and unlabeled data to further tune the pre-trained model.}}\label{fig:framework}\vspace{-10pt}
	 \end{figure}
	 
	%by leveraging the benefit of unlabeled data using contrastive learning and consistency regularization.
	In this paper, we propose a framework, displayed in Fig.~\ref{fig:framework}, to integrate the state-of-the-art active learning and self-/semi-supervised learning algorithms to build accurate models in the very few labeled data regime. Our framework enables us to isolate and assess the benefits of individual components which are evaluated using state-of-the-art methods (e.g.~Core-Set, VAAL, simCLR, FixMatch). Our contributions are as follows:\som{Should we discuss that we do active learning with no or few initial random labels?}
	\begin{itemize}\setlength\itemsep{0mm}
		\item To the best of our knowledge, this is the first framework that allow for integrating active learning and S4L techniques in the very few label regime.
		\item Focusing on S4L, we show that pre-training with contrastive learning has two key benefits. First, it improves the classification performance over plain semi-supervised learning. Secondly, it also stabilizes semi-supervised learning and reduces the variability when there are very few labels and the quality of the labeled examples significantly affects the outcome.
		\item Integrating S4L and active learning and using extensive experiments with contemporary methods, we demonstrate that much of the benefit of active learning is subsumed by S4L techniques. \so{Via ablation experiments, we identify the source of this phenomena as the remarkable benefit of data augmentation in improving label efficiency.}
		% and consistency regularization in semi-supervised learning by .
	\end{itemize}\vspace{-10pt}

%% file: approach.tex
\section{Proposed framework for learning with few labels}

\som{Is active labeler a phrase used by people?}
We first introduce our proposed framework shown in Fig.~\ref{fig:framework} which is composed of three main components: Self-supervised pre-training, {active-learning based labeler} and semi-supervised fine-tuning. We first train a self-supervised model as our pre-trained model on fully unlabeled data.Then we combine the unlabeled data and pre-trained model as input of the active labeler with a certain labeling budget. The active labeler selects the input examples for labeling by actively selecting the most informative examples in unlabeled dataset. This way the original unlabeled dataset becomes partially labeled and is composed of labeled and unlabeled components. Finally, a semi-supervised learning algorithm uses the partially-labeled dataset and the pre-trained model for training classifiers which jointly takes advantage of the labeled and unlabeled data. We shall ablate these three main components to study the benefit of each of them. We discuss our setup and ablations on each component as follows.

\noindent{\textbf{Self-supervised pre-training:}}  Our self-supervised pre-training uses Debiased Contrastive Learning (DCL)~\cite{chuang2020debiased} which improves over simCLR~\cite{chen2020simple} by effectively removing false negative pairs in contrastive loss. In our comparison experiments, we choose either random initialization or DCL pre-training.

\noindent{\textbf{Active learning: }} Besides random sampling which randomly selects the examples for labeling, we use two mainstream active learning algorithms in our experiments: Core-Set~\cite{sener2017active} and Variational Adversarial Active Learning (VAAL)~\cite{sinha2019variational}. Core-Set selects the examples to label (under a budget constraint) to achieve maximum $\delta$-radius coverage over all inputs in the latent space. VAAL trains a discriminator to identify whether the representation of an example comes from labeled set or unlabeled set while a Variational Autoencoder (VAE) tries to generate representations which is not distinguishable for the discriminator. VAAL selects examples with the lowest discriminator certainty as the most informative examples.

% at the same time
\noindent{\textbf{Semi-supervised learning: }} For this part, we mainly use FixMatch which yields state-of-the-art results.  FixMatch minimizes a combination of the standard labeled loss and a consistency loss between the weak and strong augmentation of the unlabeled data. FixMatch can achieve near supervised performance with few labels on CIFAR10 dataset. We additionally study (i) plain pseudo-labeling (a.k.a.~self-training) without augmentation and (ii) supervised training (which does not utilize unlabeled data) to further explore the impact of data augmentation and pseudo-labeling.\vspace{-5pt}

%% file: experiments.tex
\begin{table*}[]
	\centering
	\resizebox{\textwidth}{!}{%
		\begin{tabular}{@{}ccccccccc@{}}
			\toprule
			& \multicolumn{4}{c}{Cifar 10} & \multicolumn{4}{c}{Cifar 100} \\ 
			\midrule
			\multicolumn{1}{c}{Methods} &40 Labels & 150 Labels & 250 Labels& Supervised & \multicolumn{1}{|c} {400} Labels & 1500 Labels & 2500 Labels & Supervised \\ 
			\midrule
			\multicolumn{1}{c|}{None + Random sampling + Pseudo-labeling} & 18.73 & 38.79 & 41.80 & 93.32 &  \multicolumn{1}{|c}{} 10.97  & 29.22  & 38.05 & 75.94  \\
			\multicolumn{1}{c|}{None + Random sampling + FixMatch} & 86.19 & 92.86 & 93.08 & 93.32 & \multicolumn{1}{|c}{} 40.08 & 61.11 & 67.31 &  75.94 \\
			\midrule
			\multicolumn{1}{c|}{DCL + Random sampling + Pseudo-labeling} & 70.77 & 75.11 & 78.83 & 94.67  & \multicolumn{1}{|c}{} 28.62 & 43.24 & 49.93  & 76.87  \\
			%\multicolumn{1}{c|}{MoCo + Random sampling + Pseudo-labeling}  &  &  &  &  & \multicolumn{1}{|c}{}  &  &   &  \\
			\multicolumn{1}{c|}{DCL + Random sampling + FixMatch} & 93.23 & 93.90 & 94.90 & 94.67 & \multicolumn{1}{|c}{} 47.66 & 65.25 &  67.49 & 76.87  \\
			%\multicolumn{1}{c|}{MoCo + Random sampling + FixMatch} &  &  &  &  & \multicolumn{1}{|c}{}  &  &   &  \\
			\bottomrule
		\end{tabular}
	}\caption{The ablation study of applying DCL pre-training shows that constrastive learning pre-training boosts the semi-supervised learning performance especially when there are very few labeled examples. Furthermore, the contrastive learning pre-training also stabilizes the semi-supervised performance when there are 40/400 labels in CIFAR10/CIFAR100. In this regime the quality of the chosen labeled examples significantly affects training especially when there's no pre-trained model (upper block). }\label{table:ablation_pre_train}\vspace{-13pt}
\end{table*}

\section{Numerical Experiments}
\subsection{Experimental Setup}
In experiments, we train the WideResNet-2-28 model \cite{zagoruyko2016wide} which is widely used in semi-supervised learning research~\cite{berthelot2019mixmatch,berthelot2019remixmatch,sohn2020fixmatch} on CIFAR10 and CIFAR100 datasets. Below we introduce the detailed setup for the state-of-the-art methods we use for the evaluations within our framework.

   \noindent\textbf{DCL:} We train DCL for 500 epochs with projection head of 128 dimensions (same as the feature dimension). Two separate runs of different random seeds are conducted and averaged to reduce the randomness of our results.
	
	\noindent\textbf{VAAL and CoreSet:} The feature representations of DCL pre-trained models are used as input for both VAAL and Core-Set. Initially, we pick 3 samples for each class except the 40 samples case in CIFAR10 where we pick 2 samples for each class initially to avoid the extreme case where some classes are not sampled at all, especially for the very few label setup. To dive into details, in VAAL, we use the features before ResNet's global pooling layer and also modify the Variational Autoencoder (VAE) of VAAL to the same dimension as the input feature. The final samples to be labeled are obtained at the end of 50 epochs of training. For the Core-Set approach, we use the features of ResNet after global pooling and reduce their dimension via principal component analysis (PCA). 
	
	\noindent\textbf{Fixmatch and Pseudo Labeling:} In both FixMatch and Pseudo-labeling experiments, we train the model for 1500 epochs (approximately $180,000$ batch updates) and report the test accuracy at the end of training. For the labeled-only experiment, we report the best test accuracy during training because the model easily overfits when there are few labels which results in the deterioration of the test accuracy.
	
	With these experimental setup, below we discuss the main outcomes of our experiments.

	\subsection{Self-supervised learning improves over semi-supervised learning (SSL)}
	
	In this section, we inspect the advantage of self-supervised pre-training through Table~\ref{table:ablation_pre_train}. Comparing the first and third lines where the model trained with random sampling and Pseudo-labeling, the DCL pre-training provides a significant boost compared to random initialization. On the other hand, observe that the model in the second line trained with FixMatch with random initialization already achieves promising performance. The DCL pre-training still improves the performance on top of FixMach, especially when there are very few labels.  That further emphasize the importance of pre-trained models which build a good feature representation and stabilize the semi-supervised performance in the very few label scenario. 
	\begin{table*}[]
		\centering
		\resizebox{\textwidth}{!}{%
			\begin{tabular}{@{}ccccccccc@{}}
				\toprule
				& \multicolumn{4}{c}{Cifar 10} & \multicolumn{4}{c}{Cifar 100} \\ 
				\midrule
				
				\multicolumn{1}{c}{Methods} &40 Labels & 150 Labels & 250 Labels& Supervised & \multicolumn{1}{|c} {400} Labels & 1500 Labels & 2500 Labels & Supervised \\ 
				\midrule
				\multicolumn{1}{c|}{None + Random sampling + Labeled only} & 15.01 & 32.89 & 40.85 &  93.32 &  \multicolumn{1}{|c}{} 9.75  & 23.25 & 33.24 & 75.94   \\
				\multicolumn{1}{c|}{None + Core-set + Labeled only} &  20.01 &  34.02 & 42.97 & 93.32 & \multicolumn{1}{|c}{} 11.87 & 26.49 & 36.01 &  75.94  \\
				\multicolumn{1}{c|}{None + VAAL + Labeled only} & 19.23 & 35.48 & 44.40 & 93.32 & \multicolumn{1}{|c}{}  10.57& 23.64 & 34.23 &  75.94  \\
				\midrule
				\multicolumn{1}{c|}{DCL + Random sampling + Labeled only} & 65.63 & 74.80 & 76.73 &  94.67 &  \multicolumn{1}{|c}{}  29.65 & 42.63 & 49.14 & 76.87   \\
				%\multicolumn{1}{c|}{MoCo + Random sampling + Labeled only} &  &  & & & \multicolumn{1}{|c}{}  & & &   \\
				\multicolumn{1}{c|}{DCL + Core-set + Labeled only} &  67.97 &  74.52 & 77.35 & 94.67 & \multicolumn{1}{|c}{} 30.19 & 43.11 & 49.70 &  76.87  \\
				\multicolumn{1}{c|}{DCL + VAAL + Labeled only} & 67.03 & 75.82 & 77.46& 94.67 & \multicolumn{1}{|c}{} 28.57 & 43.54 & 50.29 &  76.87  \\
				%\multicolumn{1}{c|}{MoCo + Core-set + Labeled only} &  &  & & & \multicolumn{1}{|c}{}  & & &   \\
				%\multicolumn{1}{c|}{MoCo + VAAL + Labeled only} &  &  & & & \multicolumn{1}{|c}{}  & & &   \\
				\midrule
				\multicolumn{1}{c|}{None + Random sampling + Pseudo-labeling}  & 18.73 & 38.79 & 41.80 & 93.32 &  \multicolumn{1}{|c}{} 10.97  & 24.94  & 36.50 & 75.94 \\
				\multicolumn{1}{c|}{None + Core-set + Pseudo-labeling}  &23.06  &37.98    & 42.26 & 93.32 & \multicolumn{1}{|c}{}   9.06 & 26.63 &  36.63 & 75.94 \\
				\multicolumn{1}{c|}{None + VAAL + Pseudo-labeling} & 23.17 & 38.21 & 41.82 & 93.32  & \multicolumn{1}{|c}{} 9.96 & 26.19 & 37.40  & 75.94 \\
				\midrule
				\multicolumn{1}{c|}{None + Random sampling + FixMatch}  & 86.19 & 92.86 & 93.08 & 93.32 & \multicolumn{1}{|c}{} 40.08 & 61.11 & 67.31 &  75.94 \\
				\multicolumn{1}{c|}{None + Core-set + FixMatch}  &   86.01 &  92.14 & 93.21  & 93.32 & \multicolumn{1}{|c}{} 41.61 & 60.79 & 67.15  & 75.94 \\
				\multicolumn{1}{c|}{None + VAAL + FixMatch} &  86.94 & 92.99 & 93.02 & 93.32  & \multicolumn{1}{|c}{} 40.33 & 61.10 &  67.63 & 75.94 \\
				\bottomrule
			\end{tabular}
		}\caption{In this table, we provide further study on various combinations of self-/semi-supervised and active learning algorithms to investigate the impact of individual components on the accuacy and which component subsumes the benefit of active learning.}\label{table:ablation_who}\vspace{-13pt}
	\end{table*}
	\subsection{Marginal Benefit of Active Learning}
	
	We observe that in Table~\ref{table:ablation_active}, active learning failed to provide improvement when we combine active learning with state-of-the-art S4L techniques. That raises the question that \emph{which component(s) in the framework undermine and subsume the benefit of active learning?}. 
	
	In the following discussion, we propose the explanation that advanced data augmentation and consistency regularization enhance the label efficiency by utilizing the unlabeled data and reduce the contribution of active learning. Extensive experiments in Table~\ref{table:ablation_who} provide further investigations on the impact of data augmentation.
	
	\noindent\textbf{What Subsumes the Benefit of Active Learning:}
	In Table~\ref{table:ablation_who}, we conduct a detailed ablation study on each component in the framework. In the first and third blocks, the model trained with the random initialization and with Labeled-only or Pseudo- labeling achieves a noticeable performance boost when ap- plying active learning techniques. We remark that the above methods only employ standard data augmentation, such as random crop and horizontal flip. However, in the second block where DCL pre-training is applied, the benefit of active learning gradually diminishes. The fourth block shows that FixMatch achieves near supervised performance with very few labels while the improvement from active learning almost entirely disappears.
	Differing from the methods in the first and third blocks, both DCL and Fixmatch exploit the unlabeled data by stronger data augmentation and consistency regularization. DCL uses a bit weaker data augmentation than FixMatch and minimizes the distance of two data augmentations of the same sample in the feature space while maximizing the distance of different samples. FixMatch employs stronger data augmentation tech- niques and enforces the consistency between weak and strong augmentations of the same data. Observe that although the benefit of active learning is marginal in DCL pre-training, it still noticeable. Thus we conclude that the data augmentation subsumes the benefit of active learning and the improvement due to active learning depends on the strength of data augmen- tation. We also remark that consistency regularization plays an essential role in enhancing the label efficiency by making the examples of the same class closer in the feature space.\vspace{-10pt}

%	The two main suspects are per-train model and SSL algorithm. Thus, we provided Table.~\ref{table:ablation_who} the information of the combination of pre-train / none pre-train , without / with active learning, Fixmatch(stronger SSL) / Pesudo Labeling(weaker SSL) / label only to show the final results of ablation analysis. 
%	
%	From the last block of top-down order in Table.~\ref{table:ablation_who}, when Fixmatch appeared without pretrain model, the result is close to fully supervised learning. However, if we used the DCL without FixmatchL as the second block, the result is not that well.  Also, if we put the first blocks of Table.~\ref{table:ablation_active} and the second blocks of Table.~\ref{table:ablation_pre_train} together, DCL benefits really marginal when the FixMatch (strong SSL) in the pipeline. Thus, we can infer that FixMatch includes most of the benefits and make other components benefit less than expectation. However, Pesudo Labeling (weaker SSL) is another story, it is weaker than DCL compared to its fully supervised learning, so by proportion of including benefit is Fixmatch (stronger SSL) >> DCL $>$ esudo Labeling (weaker SSL).
%	
%	Although the result is not that significantly be boosted by active learning algorithm, we can still see the improvement of active learning from the top block comparison between with and without active learning algorithm in the none pre-train label only scenario. 
%   

%% file: related.tex
\section{Prior Art}
	\noindent{\textbf{Self-Supervised Learning: }}The basics of recent success on contrastive visual representation learning in self-supervised learning can be traced back to \cite{hadsell2006dimensionality} where they learn representations by contrasting positive pairs against negative pairs. Along this line, \cite{wu2018unsupervised,tian2019contrastive} use memory banks to store the instance class representation vector. \cite{chen2020big,he2020momentum,chuang2020debiased} explores the role of momentum and large batch size in contrastive learning. Especially, \cite{chen2020big} mentions that by using supervised fine-tuning on 10\% labeled data along with model distillation, the distilled model achieves the same performance as fully supervised learning. \cite{zoph2020rethinking} discusses how self-training outperforms normal pre-training in terms of stronger augmentation. However, there's a lack of research on exploring the benefit of contrastive learning pre-train where the feature representations of the same class are typically clustered and correlated. To the best of our knowledge, we are the first paper to provide the framework combining S4L and active learning to discuss how the contrastive learning pre-train stabilizes semi-supervised learning performance on very few labeled scenarios.

	\noindent{\textbf{Active Learning: }}In active learning, we mainly focus on pool-based active learning where algorithms use different sampling strategies to determine how to select the most informative samples. \cite{sener2017active,dutt2016active} select the most informative samples by maximizing the diversity in a given batch. \cite{sinha2019variational,mottaghi2019adversarial,kim2020task} pick samples by maximizing adversarial or variational autoencoder  (VAE) loss. However, none of the above methods consider their algorithms in a regime where the labeling budget is extremely small, for example, labeling lower than 10\% or even 1\% of total training data. Moreover, some state-of-the-art semi-supervised algorithm outperforms active learning with less labeled data by utilizing data augmentation. We will study the impact of a small budget in active learning when combined with S4L algorithms that take advantage of unlabeled data.
	
	\noindent{\textbf{Semi-Supervised Learning: }}Semi-supervised learning (SSL) methods improve the model's performance by leveraging the unlabeled data. State-of-the-art SSL methods~\cite{berthelot2019remixmatch,sohn2020fixmatch,smith2020building} mostly use pseudo labeling and consistency regularization when training with unlabeled data to regularize the training. Recently, FixMatch~\cite{sohn2020fixmatch} achieves supervised accuracy on CIFAR10 dataset with a total of 250 labeled samples by exploiting the benefit of strong data augmentation. However, it is sensitive to hyper-parameter selection and requires heavy hyper-parameter tuning. Moreover, in the case of less labeled samples, since the samples are randomly selected, the performance is largely depends on the quality of the labeled samples. In our paper, we show that training on constrastive learning pre-train stabilizes the variation when selecting low quality labeled data and further improves performance.
	
	%Semi-supervised learning (SSL) methods improve the model's performance by leveraging the unlabeled data. State-of-the-art SSL methods~\cite{berthelot2019remixmatch,sohn2020fixmatch,smith2020building} mostly use pseudo labeling and consistency regularization when training with unlabeled data to regularize the training. Recently, FixMatch~\cite{sohn2020fixmatch} achieves supervised accuracy on CIFAR10 dataset with a total of 250 labeled samples. However, it may requires heavy hyper-parameter tuning. Moreover, in the case of less labeled samples, since the samples are randomly selected, the performance is largely depends on the quality of the labeled samples. In our paper, we show that training on constrastive learning pre-train stabilizes the variation when selecting low quality labeled samples and further improves performance.

%% file: conclusion.tex
\section{Discussion and Future Work}
% the performance it is not that satisfied, 
This work proposed a new framework integrating active learning with S4L. Our extensive experiments demonstrate that self-supervised learning and semi-supervised learning interact synergistically, and each of them is a powerful component that drastically reduces the required labeling. In contrast, we evaluated two popular active learning algorithms (CoreSet and VAAL) and showed that the benefit of active learning on top of the state-of-the-art S4L methods is surprisingly marginal.

%the two powerful algorithms, self-supervised learning and SSL works sequential with few label scenario makes the improvement of whole pipeline really marginal. However, we can learn that self-supervised and SSL algorithm is really powerful.

As future work, a clear direction is designing better active learning algorithms that work well in the very few label regime. The existing active learning algorithms require too many labels to perform well and at that stage S4L already reaches very high accuracy while subsuming the benefits of active learning. For computer vision benchmarks, the ideal active learning algorithms should be able to select the most informative 1 to 10 examples per class out of thousands to be able to establish improvement over S4L. This would require fundamentally different algorithms that are much more precise at selecting the most informative examples. A related question is whether it is (theoretically) unavoidable that active learning is less beneficial when one can extract diverse information from the unlabeled data using extensive data augmentation.

%fundamental rethinking of current tec

%\so{The ideal active learning algorithm should not suffer from imbalanced class labeling and should have the ability to select the examples that can represent other classes corresponding to representation or to the pre-trained model. We believe that nowadays contrastive learning based self-supervised learning model is better to work with active learning algorithms dynamically.}